%% file: acl_latex.tex
\pdfoutput=1

\documentclass[11pt]{article}

\usepackage{acl}
\usepackage{multicol}
\usepackage{bm}
\usepackage{amsmath}
\usepackage{amssymb}
\usepackage{multirow}
\usepackage{graphicx}
\usepackage{times}
\usepackage{soul}
\usepackage{tablefootnote}
\usepackage[normalem]{ulem}
\usepackage{multirow}
\usepackage{makecell}
\usepackage{enumitem}
\usepackage{adjustbox}
\usepackage{amsmath}
\usepackage{array}
\usepackage{graphicx}
\usepackage{comment}
\usepackage{booktabs}
\usepackage{latexsym}
\usepackage{bigstrut}
\usepackage{booktabs} 
\usepackage{xcolor}
\usepackage{cleveref}

\crefname{section}{§}{§§}
\usepackage[T1]{fontenc}
\input{definition}
\usepackage[utf8]{inputenc}

\usepackage{microtype}

\title{Disentangling Reasoning Capabilities from Language Models with Compositional Reasoning Transformers}

\author{Wanjun Zhong$^{1}$\thanks{\ \ \ Equal contributions during internship at Microsoft Research Asia.}, Tingting Ma$^{2*}$, \textbf{Jiahai Wang$^1$, Jian Yin$^1$}, \\\textbf{Tiejun Zhao$^{2}$}, 
\textbf{Chin-Yew Lin$^3$ and Nan Duan$^3$
} \\
	$^1$ Sun Yat-sen University \quad $^2$ Harbin Institute of Technology \\
	$^3$ Microsoft Research Asia \\
	{\tt zhongwj25@mail2.sysu.edu.cn, hittingtingma@gmail.com} \\
    {\tt \{wangjiah,issjyin\}@mail.sysu.edu.cn, tjzhao@hit.edu.cn}\\
	{\tt \{cyl, nanduan\}@microsoft.com}; \tt 
}

\begin{document}
\maketitle
\begin{abstract}
\input{chapters/0-abstract}
\end{abstract}
\input{chapters/1-introduction}
\input{chapters/3-methods}

\input{chapters/4-framework}

\input{chapters/5-experiments}
\input{chapters/2-related-work}

\input{chapters/6-discussions}
\input{chapters/7-Conclusion}
\bibliography{anthology,custom}
\bibliographystyle{acl_natbib}

\appendix

\input{chapters/appendix}

\end{document}

%% file: definition.tex
\newcommand{\modelname}{{\usefont{T1}{ppl}{m}{n}ReasonFormer}}

%% file: chapters/0-abstract.tex
This paper presents \modelname, a unified reasoning framework for mirroring the modular and compositional reasoning process of humans in complex decision-making.
Inspired by dual-process theory in cognitive science, the representation module (automatic thinking) and reasoning modules (controlled thinking) are decoupled to capture different levels of cognition. 
Upon the top of the representation module, the pre-trained reasoning modules are modular and professional in specific and fundamental reasoning skills (e.g., logic, simple QA, etc).
To mimic the controlled compositional thinking process, different reasoning modules are dynamically activated and composed in both parallel and cascaded manners to control what reasoning skills are activated and how deep the reasoning process will be reached to solve the current problems. 
The unified reasoning framework solves multiple tasks with a single model, and is trained and inferred in an end-to-end manner. 
Evaluated on 11 datasets requiring different reasoning skills and complexity, \modelname~demonstrates substantial performance boosts, revealing the compositional reasoning ability. 
Few-shot experiments exhibit better generalization ability by learning to compose pre-trained skills for new tasks with limited data, and decoupling the representation module and the reasoning modules. 
Further analysis shows the modularity of reasoning modules as different tasks activate distinct reasoning skills at different reasoning depths. 

%% file: chapters/1-introduction.tex
\section{Introduction}
\begin{figure}[t]
	\centering
	\includegraphics[width=0.48\textwidth]{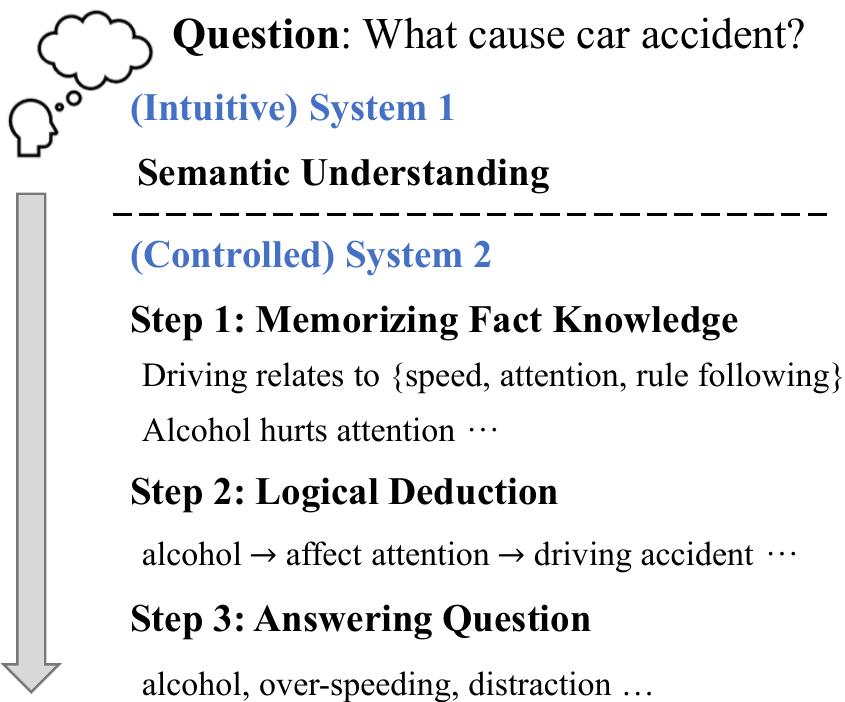}
	\caption{
	Compositional reasoning process of humans in complex decision-making. Humans solve the problems by cascaded executions of fundamental skills.}
\label{fig:example}
\end{figure}
Prevailing language models (LMs) \cite{bert,gpt-3} demonstrate impressive performance in natural language processing tasks, and have ushered in a new trend in AI research.
Despite the emerging fervor, the homogeneous LMs relying on a single call of the model are less modular and are hard to explicitly model the complex reasoning process~\cite{shallow-reasoning} like humans.

In the dual-process theory \cite{thinking-fast-slow} in cognitive psychology, there are two cognitive systems interacted to form a whole reasoning process. 
System 1 (automatic thinking) generates intuitive patterns of ideas, and System 2 (controlled thinking) constructs reasoning in an orderly logical series of compositional reasoning processes. 
Besides, in the process of System 2, different functional brain areas could be modular and interact with each other. 
System 2 can decide how to compose different reasoning skills and when to stop thinking.
As the example shown in Fig. \ref{fig:example}, when finding the cause of a car accident, humans intuitively comprehend the question (System 1), and then conduct compositional reasoning (System 2: recalling fact $\rightarrow$ logical deduction $\rightarrow$ answering question).

We would like to incorporate this mechanism into AI models in decision-making,
and make the following assumptions: (1) the representation module (System 1) and reasoning module (System 2) can be decoupled and (2) the ``complicated" reasoning process can be disentangled into multi-step executions of compositional ``fundamental" reasoning modules, whose compositionality can be learnt with limited data. 
Also, the ``fundamental" nature of basic reasoning skills allows them to have rich training instances for reliable skill pre-training.

Under these motivations, this paper proposes the modular and compositional reasoning framework - \modelname, to mirror human's compositional reasoning process, with the following characteristics: (1) the representation module and reasoning modules are decoupled; (2) reasoning modules are modular and professional in fundamental reasoning skills; 
(3) reasoning modules are compositional in parallel and cascaded manner, to dynamically decide the activated reasoning skills and the reasoning complexity;
(4) the general-purpose reasoning framework is end-to-end and unified in solving multiple tasks with one model.

Specifically, the representation module learns contextual representations of problems. 
Upon the top of the it, there are cascaded reasoning modules to perform compositional multi-step reasoning. 
The reasoning modules are pre-trained to expert in specific reasoning skills (e.g., logic, QA, fact, etc.). 
These pre-trained reasoning skills are considered relatively fundamental and have rich resources.
Two additional blocks complete the whole framework: the reasoning router and the reasoning adapter. The reasoning router decides which reasoning skills are activated in each reasoning step, and when to stop the reasoning process. 
The adapter adapts the reused reasoning modules to different steps of the reasoning process. 

We comprehensively evaluate the framework on 11 datasets emphasizing different reasoning skills and complexity, and highlight the following findings:
(1) Substantial performance boosts demonstrate models' harvest of compositional reasoning ability, and both the reasoning-centric pre-training and reasoning adapter bring compounding performance gains.
(2) Results of few-shot experiments show that specialized modules enables better generalization by learning to compose pre-trained skills for low-resource tasks, and decoupling of representation module and reasoning modules.
(3) Further analysis reveals the distinct reasoning skills required for different tasks at different reasoning depths, shoring up the modularity of reasoning modules.

%% file: chapters/3-methods.tex
\section{Reasoning Skills Formulation}
\label{sec:skill-selection}
\begin{figure*}[t]
	\centering
	\includegraphics[width=0.9\textwidth]{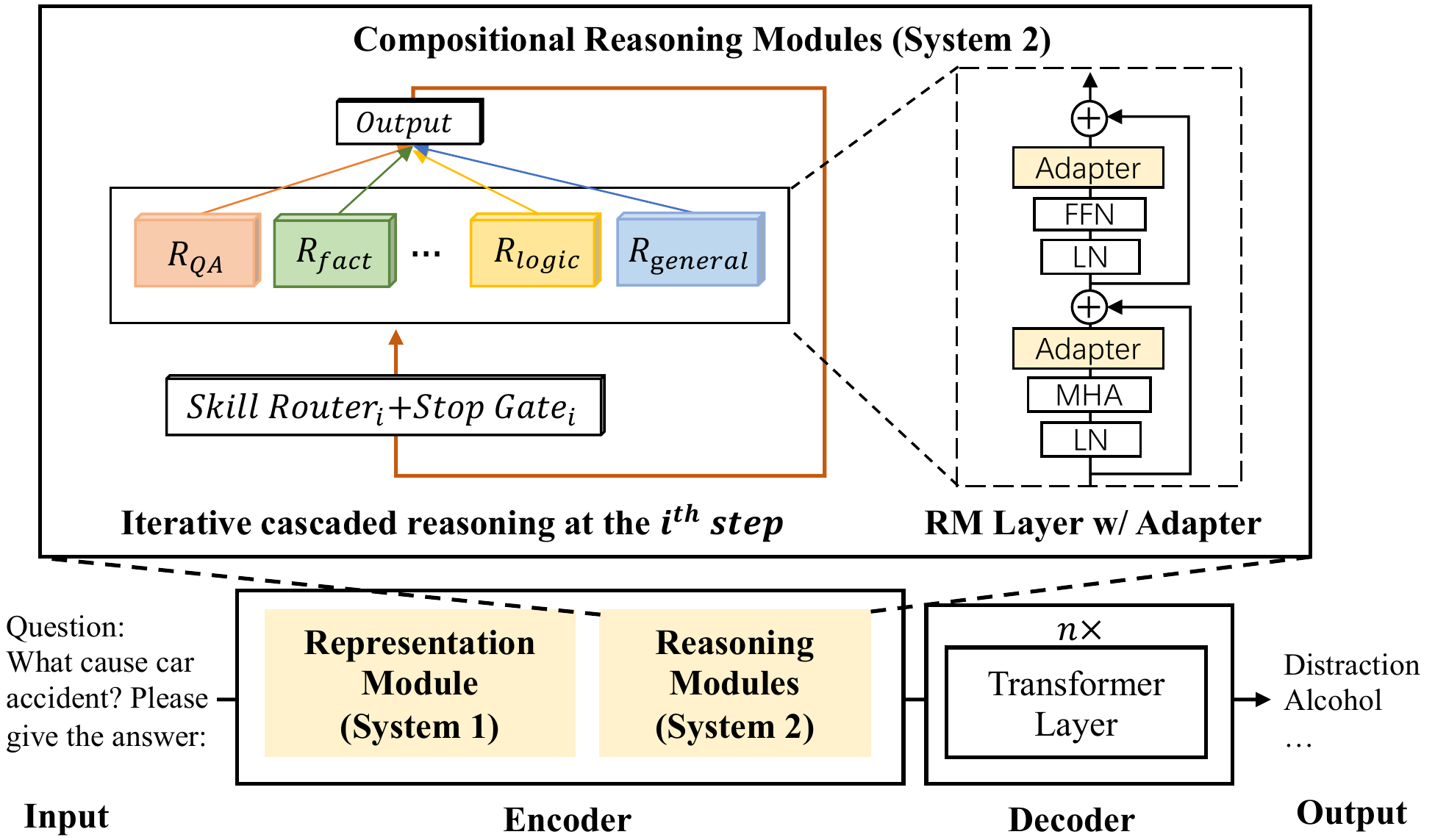}
	\caption{\modelname~framework. The representation module (\cref{sec:chap04:representation-module}) and reasoning modules (RMs) (\cref{sec:chap04:reasoning-module}) are decoupled to form the compositional reasoning process. The RMs are pre-trained with different reasoning skills $R_{skill}$ (\cref{sec:skill-selection}). 
	The reasoning adapter (\cref{sec:chap04:shared-module}) adapts the shared RMs to different reasoning steps. Router decides activated skills. Stop gate decides when to stop reasoning (\cref{sec:chap04:reasoning-router}).
	Red lines indicate cascaded reasoning process. }
\label{fig:pipeline}
\end{figure*}
The compositional reasoning process of LMs' relies on the pre-training of several fundamental reasoning skills and their compositionality. Hence, the selection of skills is critical.
\paragraph{Selection Principles.}
There are two major principles in selecting skills: 
(1) \textbf{Fundamental}: 
Complex problems can be decomposed and solved by simpler basic skills. 
So the basic skills should be more fundamental, well-defined, and can be covered in the required skill set of as many tasks as possible; 
(2) \textbf{Resourceful}:
Reliable skill pre-training requires large-scale pre-training data. 
However, in the real-world scenario, the annotated data is expensive to obtain for most reasoning tasks.
So it is expected that there are already rich resource or data can be collected via self(semi)-supervised manner.
\paragraph{Basic Skills Selection.}
Humans always solve complex problem with fundamental skills, like understanding key information (e.g., entity and its type) of events, recalling related facts, understanding causal relations between events, and extracting answers for the question. 
This motivates us to select the following basic skills: 
the \textbf{logic ability} to logically deduce the cause or consequence of events; 
\textbf{simple question answering (QA)} to understand the context and answer simple questions;
\textbf{named entity recognition (NER)} to identify important entities in the context; 
\textbf{natural language inference (NLI)} to identify semantic relevance of two sentences and
\textbf{factual knowledge} to memorize commonsense knowledge and understand daily events. 
There is an additional \textbf{general} skill to learn the commonly shared knowledge across selected skills.
We keep this setting in our paper as they are relatively well defined and resourceful
\footnote{It is worth noting that this selection is tentative. There are plausible ways for selecting basic skills or knowledge domains, which also inspire future directions. }. 

We adopt self-supervised methods to construct pre-training corpus for \{\textit{logic ability}, \textit{factual knowledge}, \textit{NER}\}, semi-supervised method to construct pre-training corpus for \textit{simple QA}, and large-scale supervised data for \textit{NLI}. Further details are given in \cref{sec:pre-train-corpus} and examples are given in Appendix \ref{sec:appendix:pre-training}.


%% file: chapters/4-framework.tex
\section{\modelname~Framework}

As shown in Fig. 2, the general-purpose reasoning framework is built based on encoder-decoder architecture to process multiple tasks (i.e., all pre-training tasks and downstream tasks) with a unified model, where all tasks are tackled as unified text-to-text generation tasks. 
We first reformat all the tasks into the same format using hard prompts \cite{t0}. For example, the question-answering task input can be prompted with the template: \textit{The question is \{Question\}. Please give the answer:"}, and the expected output is the answer text.

Given the prompted task inputs, the modular and compositional framework consists of two components in its encoder: the representation module (System 1) and the reasoning modules (System 2). 

The \textbf{representation module} (\cref{sec:chap04:representation-module}) captures the intuitive understanding of problems by calculating initial contextual representations.  
Upon the top of the representation module, there are several pre-trained \textbf{reasoning modules} (\cref{sec:chap04:reasoning-module}) with different reasoning skills, waiting for interaction to form a compositional reasoning process. 
For reasoning process organization, there are \textbf{reasoning routers} (\cref{sec:chap04:reasoning-router}) to decide the (parallel) activated skills and when to stop the (cascaded) reasoning process. 

\subsection{Representation Module}
\label{sec:chap04:representation-module}
Similar to the perceptive function of System 1, the representation module targets basic contextual understanding, and builds the foundation of the following-up reasoning process.
As LMs exhibit impressive ability on contextual understanding, we build the representation module with cascaded Transformer layers. 
Given the tokenized input $X$ with length $m$, the initial representations learnt from representation module are denoted as:
\begin{equation}
    \bm{H}^{0}=\{\bm{h}^{0}_\texttt{[CLS]},\bm{h}^{0}_1,\bm{h}^{0}_2...,\bm{h}^{0}_m\}
\end{equation}
where \texttt{[CLS]} is a special token.
\subsection{Reasoning Modules}
\label{sec:chap04:reasoning-module}
To simulate the cognitive process (System 2) formed by controlled interaction between various functional areas in human brains, the reasoning modules are modular and compositional. 
Reasoning modules (\textbf{RMs}) learn different reasoning skills specified during pre-training, and are automatically composed during downstream adaptation  (\cref{sec:chap04:reasoning-module-train-adapt}) with reasoning router (\cref{sec:chap04:reasoning-router}).
Compositionality is not only at the parallel level (different skills), but also at the cascaded level (multi-step reasoning)
Since different reasoning steps intuitively model different levels of information, there are additional \textbf{reasoning adapters} to adapt the reused modules to different reasoning steps.

\subsubsection{Reasoning Modules Architecture}
\label{sec:chap04:shared-module}
Each reasoning module is implemented by several Transformer layers.
As shown in Fig.\ref{fig:pipeline}(b), the shared reasoning modules with the same skill at different reasoning depths have shared parameters (excluding the reasoning adapter). 
For example, \textit{Fact} modules at steps $\{0, 1, ... , n\}$ share major parameters.
The output from the last reasoning step will be recursively taken as the inputs of the reused reasoning modules with step-specific adapters. 
\paragraph{Reasoning Adapter.}
To adapt the reused reasoning module to different depths of the reasoning process, we add step-specific reasoning adapters to the reasoning modules. 
Inspired by \citet{adapter} on domain adaptation, as shown in Fig. \ref{fig:pipeline}, we 
add two reasoning adapters following the multi-head attention layer and FFN layer in the Transformer layer of reasoning modules. 
Besides, the reasoning adapters for different skills and different reasoning depths are non-shared.

\subsubsection{Reasoning Router}
\label{sec:chap04:reasoning-router}
To compose the reasoning process, the reasoning router is critical in deciding which skills are activated per step, and how many reasoning steps are required for problem-solving. 
As the example in Fig. \ref{fig:example}, problem-solving needs to recall facts, and make logical deductions, then answer questions. 
Therefore, the activated skills and reasoning depths may varied for every instance.

At the parallel level of each step, the \textbf{skill router} calculates activating scores for reasoning modules. 
After each reasoning step, the \textbf{stop gate} decides 
whether executed reasoning steps are sufficient in problem-solving through a stop gating mechanism.

Unlike Mixture-of-Experts (MoE) \cite{moe} that uses token-wise routing, we adopt an instance-level routing strategy, which can capture more comprehensive semantics of problems.
\paragraph{Skill Router.}
Since the $i^{th}$ reasoning step has $n$ reasoning modules: $\{R_1,\cdots,R_n\}$ and a skill router $S^i$, the output $\bm{H}^{i}$ of the $i^{th}$ reasoning step can be calculated by router-weighted averaged outputs from the $k$ activated reasoning modules:
\begin{equation}
    \bm{H}^{i} = \sum^{k}_{j=1}S^{i}(\tilde{\bm{H}}^{i-1})_{j}R_j(\tilde{\bm{H}}^{i-1})
\end{equation}
where $S^{i}(\tilde{\bm{H}}^{i-1})_{j}$ (scalar weight) and $R_j(\tilde{\bm{H}}^{i-1})$ (updated hidden vectors) are the outputs from the router and the $j^{th}$ reasoning module, respectively. 

Since deciding the skills is a non-trivial task, we adopt a relatively complex router for deeper understanding. 
We use one Transformer layer $T$ to project the original output for routing weight calculation.
Then, we use an FFN layer followed by a Softmax function for weighted score calculation:
\begin{equation}
S^{i}(\tilde{\bm{H}}^{i-1}) = \text{Softmax}(\text{FFN}(T(\tilde{\bm{H}}^{i-1})))
\end{equation}

Afterwards, we \textbf{sparsely activate} \cite{moe} $k$ reasoning modules with top-$k$ skill routing scores at each reasoning step. The router training objectives are detailed in \cref{sec:chap04:reasoning-module-train-adapt}.
\paragraph{Stop Gate.}
After each reasoning step, the stop gate decides whether the current reasoning depth is sufficient to solve the problem.
Taking $\bm{H}^{i}$ as the input, the stop gate uses a residual gating mechanism $G^i_{stop}$ to control the information flow from executed reasoning steps and calculate the final output $\tilde{\bm{H}}^{i}$ for the $i^{th}$ reasoning step by:
\begin{equation}
    \tilde{\bm{H}}^{i} = \bm{H}^{i-1} + G^{i}_{stop}(\bm{H}^{i})
\end{equation}
An FFN layer is used as the stop gate $G^i_{stop}$.
When the reasoning process is sufficient, the following-up process will be softly stopped by $G^i_{stop}$.

\subsection{Pre-training and Adaptation}
\label{sec:chap04:reasoning-module-train-adapt}
The unified model enables multi-task learning for both pre-training and downstream tasks.
The major difference between pre-training and adaptation is that only in the pre-training stage we have the supervision for the activated skills. 

\paragraph{Pre-training.} 
Before reasoning pre-training, the model weights of \modelname~ are initialized with pre-trained weights from T5 \cite{t5}.
The details of model initialization and pre-training corpus collection are introduced in \cref{sec:model-ini} and \cref{sec:pre-train-corpus}, respectively.
Since model acknowledges which skill it is learning, we add \textbf{skill routing loss} $L_r$ in addition to the teacher-forcing loss, to guide the routers in activating skills. 
For example, if the current instance focuses on logic ability, it should activate \{\textit{logic ability, general}\} skills. 
$L_r$ can be set as the cross-entropy loss for the multi-skill classification, where the activated skill has label 1 and 0 otherwise. 
During pre-training, all the reasoning steps activate the same skill for one instance.

\paragraph{Adaptation.} 
During downstream adaptation, we have no prior knowledge about the required skills for different tasks, so we expect the model can automatically learn which skills are essential for each specific task. 
Therefore, we adopt standard teacher-forcing loss for generative training.

%% file: chapters/5-experiments.tex
\section{Experiment Setup}
\input{tables/main_results.tex}

\subsection{Datasets}
To verify the effectiveness of \modelname, we extensively conduct experiments on \textbf{11} datasets emphasizing different reasoning types and complexity.
Specifically, ReClor \cite{yu2020reclor} emphasizes on logical reasoning.
Commonsense QA (CSQA) \cite{commonsenseqa}, ARC \cite{Clark2018ThinkYH}, PIQA \cite{bisk2020piqa} and HellaSwag \cite{hellaswag} stress commonsense knowledge.
Abductive NLI (aNLI) \cite{aNLI} is a natural language inference dataset.
HotpotQA \citep{DBLP:journals/corr/abs-1809-09600} and WikiHop \cite{wikihop} focus on multi-hop question answering.
MuTual \citep{DBLP:journals/corr/abs-2004-04494}, DREAM \citep{sundream2018} focus on reasoning over dialogue. 
RACE \citep{lai2017large} is a general QA dataset.
These datasets are related to the fundamental reasoning skills (\cref{sec:skill-selection}) and fit nicely for analyzing the compositional reasoning process modeled by \modelname. 

During \textbf{Evaluation}, the Hotpot QA adopts \textit{Exact Match (EM)} as the metric, while the rest tasks use \textit{accuracy} as the metric. 
The answer for multi-choice QA and classification tasks are selected by the highest log-likelihood scores of options.
\subsection{Pre-training Corpus}
\label{sec:pre-train-corpus}

To reduce the manual efforts in data collection, we mainly select self(semi)-supervised pre-training corpus construction methods. 

To improve LMs' \textbf{logic} ability, we adopt the self-supervised logical pre-training corpus built by LogiGAN \cite{pi2022logigan}, which uses logical indicators to identify logical phenomena in a general text corpus. 
For \textbf{QA-centric} pre-training, we adopt the semi-supervised pre-training corpus construction method from ProQA \cite{lewis-etal-2021-paq, zhong2022proqa}, which adopts a generation-filtering pipeline to build QA-centric corpus. 
To help the model in \textbf{identifying entities} from text, we use the self-supervised NER corpus \cite{chen2022fewner} built from Wikidata and Wikipedia anchor link.
To learn \textbf{factual knowledge}, we use Wikidata as a commonsense knowledge base to construct self-supervised pre-training corpus. Specifically, we sample 1 million fact triples from Wikidata and construct the KG completion task \cite{moiseev-2022-skill} by recovering the masked tailed entities with the head entities and relations given as inputs.

Furthermore, since \textbf{natural language inference} task already have rich supervised data, we directly use MNLI~\citep{williams-etal-2018-broad} and SNLI~\citep{bowman2015snli} datasets as the pre-training corpus. 

Finally, 1 million instances are collected for each reasoning skill, and there are 5 millions pre-training instances in total for 5 reasoning skills. 
The examples and prompts for constructing inputs/outputs of the pre-training corpus are given in Appendix \ref{sec:appendix:pre-training}.

\subsection{Models}


\subsubsection{Model Initialization}
\label{sec:model-ini}
We adopt encoder-decoder framework. In the encoder, the representation module has 9 Transformer layers, each shared reasoning module has 3 Transformer layers and the maximum reasoning depths is 3.
We initialize the major model parameters from pre-trained $\text{T5}_\text{base}$ \cite{t5}. 
Thus, the representation module is initialized by the $1^{th}\rightarrow9^{th}$ layers of T5 encoder, and the reasoning module is initialized by $9^{th}\rightarrow12^{th}$ layers of T5 encoder. 
The decoder is the same with T5.
\subsubsection{Compared Methods}
The major focuses of the experiment are to explore the effectiveness of \modelname, and verify our hypotheses that complex problems can be disentangled and solved by compositional reasoning modules, and the decoupling of representation module and reasoning modules.
We compare \modelname~with two series of methods.
\paragraph{T5 series.}
    (1) \textbf{Vanilla T5} is the released T5 model \cite{t5} (\textit{google/t5\--v1\_1\--base}) pre-trained with C4 corpus excluding other supervised data;
    (2) \textbf{Reasoning Pre-Trained T5 (RPT-T5)} is the T5 model continually pre-trained with our reasoning-centric pre-training corpus (\cref{sec:pre-train-corpus}).

 \paragraph{MoRM series.} 
    Inspired by Mixture-of-Experts (MoE) methods \cite{moe,moe-transformer}, we develop Mixture-of-Reasoning Modules (MoRM) methods for comparison. 
    Unlike MoE that builds parallel experts in the FFN layer of Transformer Layers, MoRM builds parallel reasoning modules (RMs) on the top of the representation module, and sparsely activate these RMs. 
    Specifically, after initialized with T5, the last 3 Transformer layers in the encoder are duplicated parallelly for $N_{s}$ (numbers of skills) times, and the outputs of them are weighted average by the routing scores of the activated RMs.
    It increases the model size in the similar way with \modelname, so it can verify whether the improvements are brought by the increased parameters.
    Besides, the major differences between \modelname~and MoRM are (1) MoRM involves no cascaded reasoning steps (depth=1); 
    (2) Like MoE, RMs in MoRM are jointly trained for all instances without skill routing loss (\cref{sec:chap04:reasoning-module-train-adapt}), emphasizing no expertise of RMs.
    We also report the results of MoRM after \textbf{reasoning-centric pre-training (RPT-MoRM)}. 

\input{tables/ablation.tex}
\section{Experiment Analysis}
\subsection{Main Results}

As presented in Table \ref{tab:main}, \modelname~outperform T5 series and MoRM series across all tasks emphasizing the wide scope of different reasoning skills. 
Thus, we have the following findings:
\paragraph{\modelname~> MoRM \& T5:} 
\modelname~surpasses other methods (even with more activated parameters) by a large margin, giving evidence to our primary hypothesis that the expertise of reasoning modules and the cascaded compositional reasoning process essentially help the model in solving complex reasoning problems. 

\paragraph{RPT-T5 > T5:} The substantial performance boosts brought by RPT demonstrate that reasoning-targeted pre-training is essential in injecting various reasoning abilities into LMs. 
\paragraph{Sparse v.s. Full:}
Sparse activation of RMs leads to slightly reduced but comparable performance compared with full activation.
It suggests that although activating more skills is beneficial, the most essential RM still plays the key role in problem-solving. 
The modularity of RMs can reduce the computation burden while keeping performance.

These positive findings manifest that \modelname~ can model compositional reasoning and verify our primary hypothesis that the complex problem can be decomposed and well solved with pre-trained basic skills, and the representation module can be decoupled with the reasoning modules.

\subsection{Ablation Study}
We explore the truly functional components of \modelname~through ablation studies on 7 datasets. 
We evaluate the effectiveness of the following components: (1) reasoning pre-training;  (2) cascaded reasoning mechanism; (3) expertise of reasoning skills (skill gating loss) and (4) reasoning adapter.
\input{tables/few-shot.tex}
\paragraph{Reasoning Pre-training.}
We assume that the first factor contributing to the improvements is the multi-task reasoning-centric pre-training. 
Since vanilla LMs mainly focus on learning contextual semantics, and don't emphasize higher-level reasoning ability \cite{pi2022logigan,shallow-reasoning}, it is intuitive that reasoning-driven pre-training can enhance the model in solving complex problems.
Results in Table \ref{tab:ablation} suggest that the ablation of pre-training from all models leads to a substantial performance drop, showing the importance of reasoning-centric pre-training in helping the reasoning modules to learn fundamental skills.
\paragraph{Cascaded Reasoning Mechanism.}
The second hypothesis is that the cascaded reasoning mechanism facilitates problem-solving with different complexity and composition orders. 
\textbf{Single} is an ablated version of \modelname~in which the reasoning modules are not cascaded horizontally (depth=1) and the adapter is also eliminated. 
Comparison between performances of Cascaded and Single version of \modelname~(Line 1 v.s. Line 4) demonstrates that the cascaded reasoning mechanism brings notable improvements and reveals the effectiveness of multi-step reasoning process.

\paragraph{Expertise of RMs.}
We assume that modularity and expertise of reasoning modules enables them to be flexibly composed. 
We ablate it from the Single version of \modelname ~(Line 6) by pre-training all the RMs jointly without skill routing loss (\cref{sec:chap04:reasoning-module-train-adapt}) using the whole pre-training corpus. 
The apparent performance drop suggests that the  expertise of RMs enables the model to discriminate the functionality of various skills and selectively compose them to form a whole reasoning process.
\paragraph{Reasoning Adapter.}
The reasoning adapters adapt the shared RMs to different reasoning steps. 
It is intuitively important as different levels of cognition focus on the information at different granularity.
From Table \ref{tab:ablation}, eliminating the reasoning adapter (Line 3) from \modelname~(Cascaded) harms the overall performance, testifying the distinct mechanisms at different levels of reasoning and the importance of reasoning-centric adaptation. 
\vspace{-0.15in}
\begin{figure*}[t]
	\centering
	\includegraphics[width=0.9\textwidth]{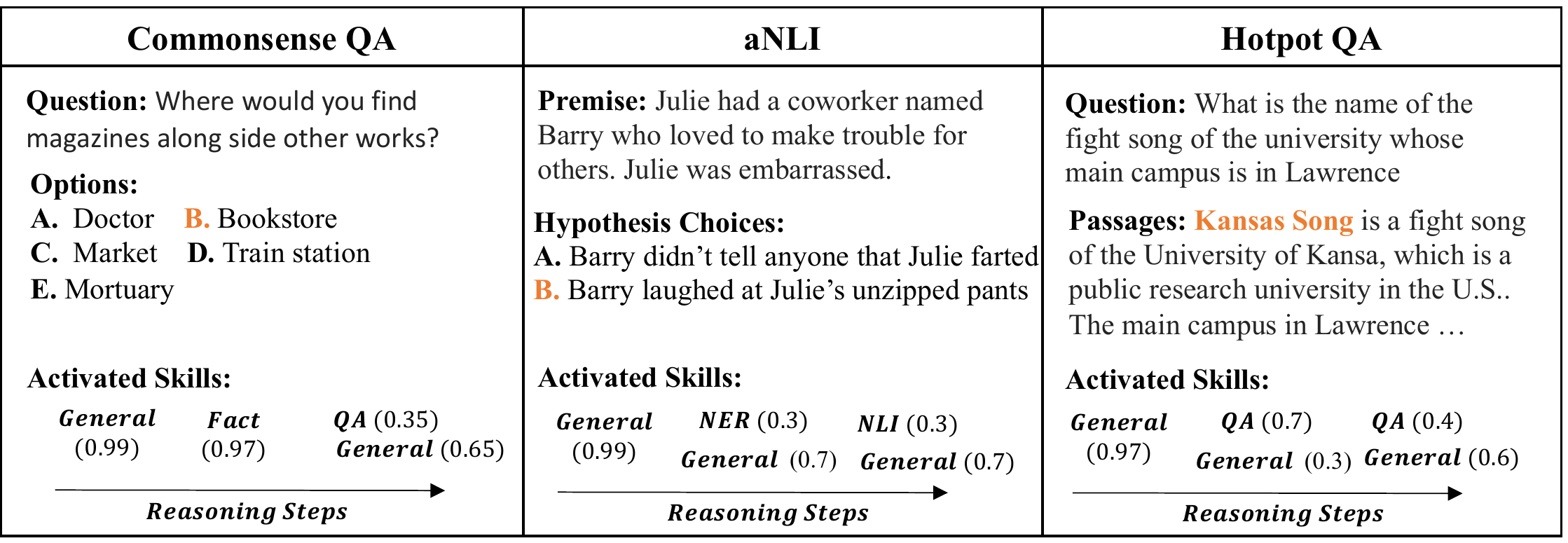}
	\caption{Case study for activated skills analysis on 3 datasets. The answer is marked with orange.}
\label{fig:moe-weight}
\vspace{-0.15in}
\end{figure*}
\subsection{Low-resource Experiments}
It is interesting to know whether the fundamental skills of RMs can be easily composed to solve new tasks with limited training data, and whether the representation module and RMs can be decoupled during adaptation. 
If the answer is true, then the model's generalization ability will be greatly enhanced with easy composition of pre-trained skills. 

Under these motivations, we conduct few-shot experiments. 
We first examine the generalization of \modelname and examine the decoupling of modules in \modelname~by freezing different modules during learning. 
We \textbf{freeze the RMs} (Line 3) to testify whether the skills can be directly reused without further fine-tuning. 
Then we \textbf{freeze the representation module} (Line 4) to verify the decoupling of representation and RMs. 
From Table \ref{tab:fewshot}, we highlight the following findings.

(1) \modelname~outperforms T5, showing that the generalization ability of \modelname~is enhanced by reasoning pre-training and explicit modeling of the compositional reasoning process.

(2) Freezing RMs (Line 3) achieves comparable and even slightly better performance than its fully tuned version, demonstrating that the learned skills can be easily composed with limited training data without further tuning RMs.


(3) Freezing the representation module (Line 4) also leads to comparable performance, proving that the representation module and RMs can be decoupled in adaptation. 
It suggests that it is feasible to reduce computation burden during few-shot adaptation by freezing the well-trained representation module and only tuning the RMs for tasks, which is more efficient when the representation module (e.g., gigantic LM) is extremely large for tuning.

(4) Freezing both modules (Line 5) hurts performance, showing that model adaptation to data distribution of specific tasks is still essential.


\subsection{Reasoning Skills Analysis}
Qualitative analyses are conducted to explore how the pre-trained skills are composed to solve different reasoning tasks, and how the skills changed at different reasoning depths. 
Therefore, we calculate the skill routing weights at every reasoning step (up to 3) for three tasks (i.e., \{\textit{Commonsense QA, aNLI, Hotpot QA}\}. 
The case study provides examples and corresponding (top 2) activated skills at each step. 
As shown in Fig. \ref{fig:moe-weight}, the activated skills are varied for different tasks, and are dynamically composed to form a series of reasoning steps.
For commonsense reasoning, it emphasizes \{\textit{fact, QA}\}. For NLI task, it emphasize \{\textit{NER, NLI}\}. 
For multi-hop QA task, it executes the QA module for multiple steps. 
The statistical analysis of averaged routing scores on the whole evaluation set also demonstrate the same trend.
These observations show improved interpretability of decision-making and give evidence to the hypothesis that the compositional cognitive process of humans can be transferred to AI model. 


%% file: tables/main_results.tex
\begin{table*}[htbp]
  \centering
  \small
  \resizebox{1.8\columnwidth}{!}{
    \begin{tabular}{cc|cc|ccc|ccc}
    \toprule
    \multirow{2}[4]{*}{Datasets} & \multirow{2}[4]{*}{Reasoning } & \multicolumn{2}{c|}{T5 Series} & \multicolumn{3}{c|}{MoRM} & \multicolumn{2}{c}{\modelname} \\
\cmidrule{3-9}          &       & T5    & RPT-T5 & w/o RPT (S) & RPT (S) & RPT (F)  & S & F \\
    \midrule
    \multicolumn{2}{c|}{Activated Paramters (M)} & 248   & 248   & 272   & 272   & 357      & 294   & 407 \\
    \midrule
    ReClor & Logic & 35.2  & 36.8  & 35.4  & 36.8  & 35.4  & 39    & \textbf{39.4} \\
    ARC   & Commonsense    & 31.4  & 32.7  & 25.4  & 34.1  & 31.1    & \textbf{35.1}  & 34.1 \\
    CSQA & Commonsense    & 56.5  & 65.1  & 57.2  & 63    & 64.7    & 66.9  & \textbf{68.2} \\
    RACE  & General  & 63.8  & 67.4  & 66.4  & 68.8  & 70.9     & 72.5  & \textbf{73.5} \\
    DREAM & General  & 59.3  & 64.5  & 56.6  & 61.8  & 67.7   & \textbf{70.5}  & \textbf{70.5} \\
    aNLI  & NLI    & 66.9  & 66.3  & 68.2  & 68.8  & 69.8   & \textbf{69.6}  & 69.5 \\
    MuTual & Dialog & 67.3  & 70.2  & 66.8  & 69.5  & 70.5    & 72.2  & \textbf{72.5} \\
    WikiHop & MultiHop & 63.6  & 66.1  & 63.5  & 66.1  & 66.9   & 67.1  & \textbf{67.4} \\
    HotpotQA & MultiHop & 61.1  & 63.3  & 63.1  & 63.3  & 63.8    & 65.2  & \textbf{65.5} \\
    Hellaswag & Commonsense    & 31.5  & 33.7  & 34.2  & 37.9  & 43   & 53.9  & \textbf{54.9} \\
    PIQA  &   Commonsense   & 61.4  & 63.3  & 64.6  & 65.4  & 67.6   & 67.5  & \textbf{67.9} \\
    \midrule
    Avg.  &       & 54.3 & 57.2 & 54.6 & 57.7 & 59.2 & 61.8 & \textbf{62.1} \\
    \bottomrule
    \end{tabular}%
    }
    \caption{Main results on 11 reasoning tasks. \textbf{RPT indicates reasoning-centric pre-training} introduced in \cref{sec:skill-selection} \& \cref{sec:pre-train-corpus}. S indicates sparse activation (top 2) of RMs, while F denotes full activation of all RMs.}
    \vspace{-0.2in}
  \label{tab:main}%
\end{table*}%

%% file: tables/ablation.tex
\begin{table*}[htbp]
  \centering
  \small
  \resizebox{1.9\columnwidth}{!}{
    \begin{tabular}{ccccccccccc}
    \toprule
    \multicolumn{1}{l}{Modules } & \multicolumn{2}{c}{Models/Dataset } & ARC   & CSQA  & DREAM & WikiHop & HotpotQA & Hellaswag & PIQA  & Avg. \\
    \midrule
    \multirow{3}[2]{*}{Cascaded (S)} & \multicolumn{2}{c}{\modelname} & 35.1  & 66.9  & 70.5  & 67.1  & 65.2  & 53.9  & 67.5  & 60.9 \\
          & \multicolumn{2}{c}{w/o RPT} & 24.1  & 56.9  & 59.5  & 64.2  & 64.4  & 34.7  & 65.6  & 52.8 \\
          & \multicolumn{2}{c}{w/o adapter} & 33.1  & 64.2  & 68.4  & 66.8  & 64.8  & 47.1  & 67.4  & 58.8 \\
    \midrule
    \multirow{3}[2]{*}{Single (S)} & \multicolumn{2}{c}{\modelname} & 34.8  & 63.7  & 65.9  & 66.6  & 63.9  & 39.7  & 66.3  & 57.3 \\
          & \multicolumn{2}{c}{w/o RPT} & 25.4  & 57.3  & 56.7  & 63.6  & 63.1  & 34.3  & 64.6  & 52.1 \\
          & \multicolumn{2}{c}{w/o modularity} & 34.1  & 63.1  & 61.8  & 66.1  & 63.4  & 37.9  & 65.4  & 55.9 \\
    \bottomrule
    \end{tabular}%
    }
  \caption{Ablation study on 7 datasets. RPT denotes reasoning pre-training (\cref{sec:skill-selection}). 
    Modularity denotes skill gating loss (\cref{sec:chap04:reasoning-router}), and adapter is reasoning adapter (\cref{sec:chap04:shared-module}).  Cascaded and Single are different in the cascaded steps. }
  \label{tab:ablation}%
  \vspace{-0.1in}
\end{table*}%

%% file: tables/few-shot.tex
\begin{table*}[htbp]
  \centering
  \small
  \resizebox{1.8\columnwidth}{!}{
    \begin{tabular}{cc|cc|ccccccc}
    \toprule
    \multicolumn{2}{c|}{\multirow{2}[2]{*}{Models }} & Freezed  & \#Tuned & ReClor & CSQA  & RACE  & ARC   & MuTual & WikiHop & Avg. \\
    \multicolumn{2}{c|}{} & Modules & Para. (M)  & Acc   & Acc   & Acc   & Acc   & Acc   & Acc   &  \\
    \midrule
    \multicolumn{2}{c|}{T5} & no    &   248    & 28.2 & 23.2 & 26.2 & 25.1 & 32.6 & 18.3 & 25.6 \\
    \midrule
    \multicolumn{2}{c|}{\multirow{4}[2]{*}{\modelname}} & no    &   294   & 29.0    & 39.2 & 29.2 & 30.1  & 40.3 & 26.4 & 32.4 \\
    \multicolumn{2}{c|}{} & RM    &  251  & 29.4  & 39.1 & 29.2 & 31.1  & 38.7 & 26.9 & 32.4 \\
    \multicolumn{2}{c|}{} & rep.  & 230 & 29.2  & 38.9 & 29.2 & 30.1 & 38.0 & 26.5 & 32.0 \\
    \multicolumn{2}{c|}{} & RM+rep. &  188  & 29.2  & 37.8 & 28.4 & 28.8 & 31.6  & 25.4 & 30.2 \\
    \bottomrule
    \end{tabular}%
    }
  \caption{Few-shot experiments after freezing different modules. The representation module is abbreviated as rep.}
  \label{tab:fewshot}%
  \vspace{-0.15in}
\end{table*}%

%% file: chapters/2-related-work.tex
\vspace{-0.05in}
\section{Related Works}
\paragraph{Multi-step Reasoning.}
Multi-step reasoning is a characteristic of human thinking.
Multi-hop reasoning \cite{yang2018hotpotqa,yu2021multi} asks the system to logically switch attention to different contexts \cite{hybrid-chain} or make a multi-step deduction for a new conclusion \cite{propara,zhong2021ar}. 
Recently, \textit{chain-of-thought} prompting \cite{wei2022chain} provides the model with manual prompts about the intermediate reasoning steps. \citet{faithful-reasoning} use LMs to iteratively select evidence and generate inferences. 
However, they always require discrete manual-written reasoning traces.
\citet{cascade-lm} is a position paper raising interest in modeling these cascaded inference processes of LMs with a probabilistic program.
\paragraph{LM Modularity.}
Since human brains have various functional areas, it is inspiring to explore the modularity of LMs. 
Mixture-of-Experts (MoE) \cite{moe,DmitryLepikhin2020GShardSG} use experts in FFN layers for sparse learning. 
However, their major motivation is to increase the model capacity while keeping efficiency, without emphasis on the speciality of expert. 
Recent works begin to explore domain-specific experts \cite{gururangan2021demix} and modality-specific experts \cite{wang2021vlmo}.
SkillNet proposes skill-specific experts \cite{zhang2022skillnet}. 
However, the activated skills need to be manually specified, and do not explicitly model the cascaded reasoning process and disentangling of perception and cognition.

Considering these directions in the whole picture, this paper targets to explore the modeling of modular and the compositional multi-step reasoning process of AI models in an end-to-end manner.

%% file: chapters/6-discussions.tex
\section{Discussion}
Despite the positive findings, we also discuss potential limitations, challenges, and inspiring future directions to shed a light on future works.

\paragraph{Limitations.} 
Since we must select current well-defined tasks with plausible ways to collect sufficient training instances, the current selection of fundamental reasoning skills is restricted by data source and clarity of the fundamental task definition. So they may not be fundamental enough. 
For example, simple QA skill may overlap with NER skill to some extent. In the future, it is worth studying more self-supervised training task to inject more fundamental abilities into RMs. 

Furthermore, since encoder-decoder architecture is the most common choice to develop a unified model, we conduct experiments on this architecture. Next, we are curious about the effectiveness of this framework when it is adapted to encoder-only or decoder-only architectures. We will further extend this framework to different architectures and model sizes. 
\paragraph{Extension of Reasoning Skills.}
This paper demonstrates that the reasoning-centric pre-training and expertise of reasoning modules enhance the compositional reasoning ability of models. 
As mentioned in Sec. \ref{sec:skill-selection}, the selection and composition of fundamental reasoning skills have a broad space to be further explored. 
For example, numerical reasoning ability can also be selected as a new skill to solve mathematical problems. 
Furthermore, the corresponding method for skill-centric pre-training corpus construction can also be explored.  
\paragraph{Multi-domain and Multi-modal RMs.}
Knowledge domain type (e.g., finance, law) or modality type (e.g., video, image, text) can also be tackled as a skill type. 
For example, RMs can be pre-trained with knowledge from different domains and modalities, and enables knowledge composition or multi-modal understanding for problem solving.

%% file: chapters/7-Conclusion.tex
\section{Conclusion}
This paper stimulates the compositional reasoning process of humans in decision-making, and makes the following hypotheses: 
(1) the intuitive perception system (System 1) and cognitive reasoning system (System 2) can be decoupled and 
(2) the complex decision-making can be disentangled into multi-step execution of fundamental reasoning skills. 
Correspondingly, we propose \modelname, a compositional general-purpose reasoning framework. 
\modelname~decouples the representation module and reasoning modules, which are pre-trained to expert in fundamental reasoning skills.
The reasoning modules are dynamically composed in parallel and cascaded manner to form a whole reasoning process.
\modelname~is end-to-end and unified in solving multiple tasks with one model.
Extensive experiments on 11 tasks reveal the compositional reasoning ability of \modelname~and disentangling of representation and reasoning modules.

%% file: chapters/appendix.tex
\section{Example of Pre-training Tasks}
\label{sec:appendix:pre-training}
For the basic question answering skill, QA-centric pre-training uses a generation-filtering pipeline to build semi-supervised large-scale corpus \cite{lewis-etal-2021-paq, zhong2022proqa}: (1) use annotated QA data to train a passage-to-{question-answer} generator (2) taking the wikipedia passages as inputs, and generates corresponding pseudo questions and answers (3) filtering {passage, question, answer} pairs with a QA model.

For logic skill, we use the automatically constructed data from LogicGAN \cite{pi2022logigan}.It uses logical indicators (e.g., Therefore, as a result) to automatically identify logical inference phenomenon presented via natural language, and mask corresponding causes/results of events, and ask the pre-trained model to recover them to learn logical reasoning ability.

For the natural language inference, we the public annotated corpus SNLI \cite{bowman2015snli} and MNLI \cite{williams-etal-2018-broad}. Given a sentence as premise, the model is expected to predict whether the premise sentence entails the hypothesis sentence.

For the named entity recognition skill, we use weakly-annotated data \cite{chen2022fewner} obtained from Wikipedia and Wikidata. The mentions are the text with anchorlink and the types are obtained from Wikidata ``instance of'' or ``subclass of'' properties. We design three pretrain tasks similar to \citet{chen2022fewner}: 1) given the sentence, identify all mentions in the sentence 2) given the sentence and interested types, output all mentions with these types in the sentence 3) given the sentence and mentions, predict all types of the mentions.

For the fact skill, we use fact triples from Wikidata, and design a task that predict the tail entity given the head entity and relation as \citet{moiseev-2022-skill}.

\begin{figure*}[htbp]
    \centering
    \includegraphics[width=2\columnwidth]{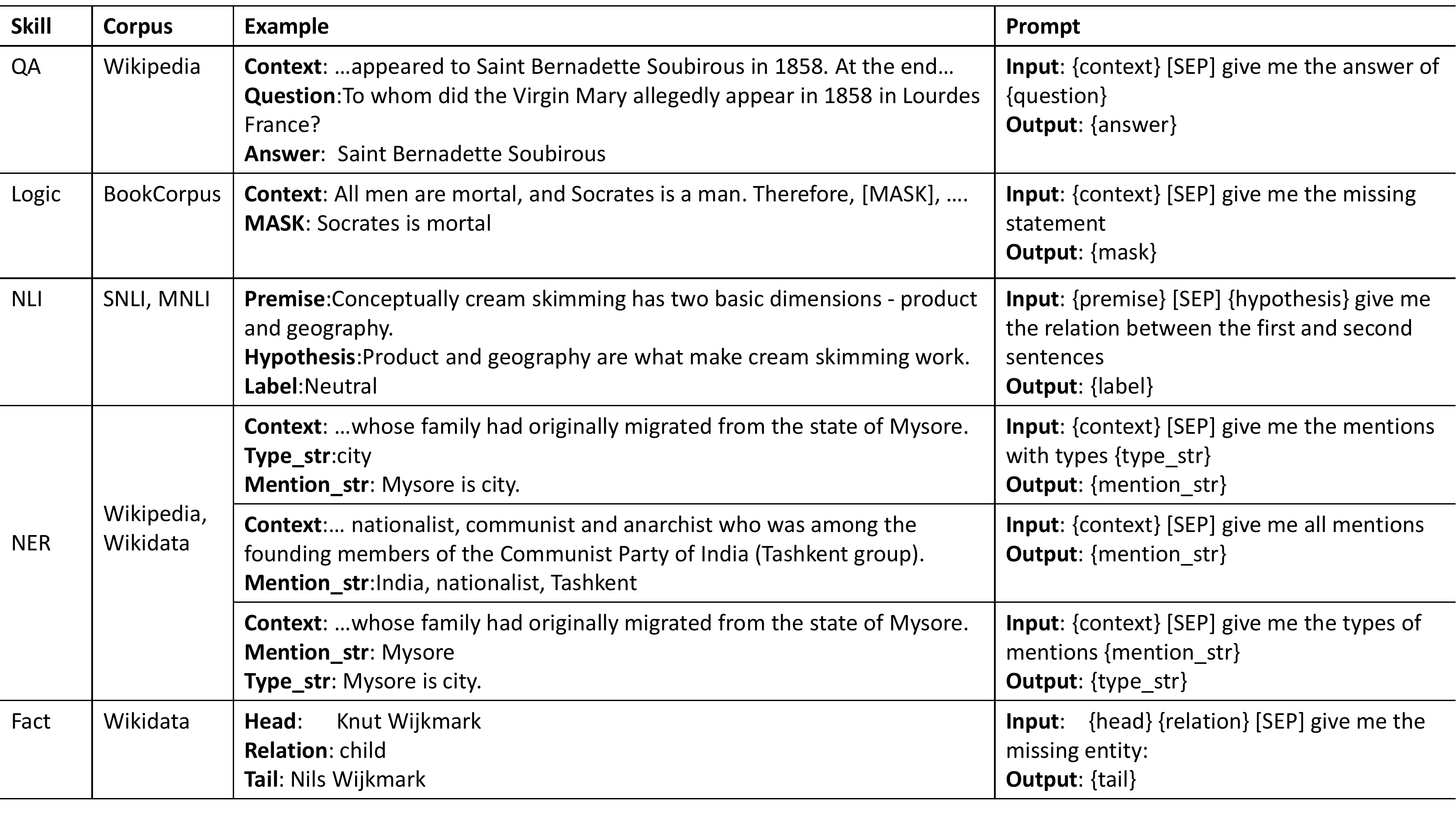}
    \caption{Examples of pretrain tasks summary.}
    \label{fig:task-example}
\end{figure*}

A summary of the examples for each tasks is presented in Fig~\ref{fig:task-example}.

\section{Implementation Details}
\label{sec:appendix:implementation}
\paragraph{Pretraining Details}
We use ``google/t5-v1\_1-base'' from HuggingFace \citep{wolf-etal-2020-transformers} implementation as base model for all our experiments. We use a learning rate of 5e-5 and train all models with 5 epochs. The warmup ratio is set to 0.1.
The total batch size is set to 72 for shared model and 64 for private model.
The down projection hidden size of adapter is set to 256. We use 8 V100 GPUs for model training.

\paragraph{Downstream Adaptation Details}
 For all the full data experiments, we use a learning rate of 1e-4 and training epoch of 10 with a batch size of 48 for our models. The model is validated at the end of each epoch. For all the few-shot experiments, we use a learning rate of 1e-5 and training epochs of 200 with a batch size of 8. The model is validated per 200 steps.